\documentclass{IOS-Book-Article}

\usepackage{mathptmx}
\usepackage{xspace}
\usepackage{graphicx}
\usepackage{amsmath}
\usepackage{algorithmicx}
\usepackage[hidelinks]{hyperref}
\usepackage{color}

\usepackage{mynatbib}
\usepackage{mytodonotes}
\usepackage{mycommands}
\usepackage{mytikz}
\usepackage{mylistings}
\usepackage{myfloat}

\def\hb{\hbox to 10.7 cm{}}

\begin{document}

\pagestyle{headings}
\def\thepage{}

\begin{frontmatter}              % The preamble begins here.

%\pretitle{Pretitle}
\title{NdFluents: A Multi-dimensional Contexts Ontology}

\markboth{}{Technical Report. September 2016.\hb}

\author[A]{\fnms{Jos\'e} M. \snm{Gim\'enez-Garc\'ia}},
\author[B]{\fnms{Antoine} \snm{Zimmermann}} and
\author[A]{\fnms{Pierre} \snm{Maret}}

%\runningauthor{B.P. Manager et al.}
\address[A]{Universit\'e de Lyon, CNRS, UMR 5516, Laboratoire Hubert-Curien, Saint-\'Etienne, France}
\address[B]{\'Ecole Nationale Sup\'erieure des Mines, FAYOL-ENSMSE,\\ Laboratoire Hubert Curien, F-42023 Saint-\'Etienne, France}

\begin{abstract}
Annotating semantic data with metadata is becoming more and more important to provide information about the statements being asserted. While initial solutions proposed a data model to represent a specific dimension of meta-information (such as time or provenance), the need for a general annotation framework which allows representing different context dimensions is needed. In this paper, we extend the 4dFluents ontology by Welty and Fikes---on associating temporal validity to statements---to any dimension of context, and discuss possible issues that multidimensional context representations have to face and how we address them.
\end{abstract}

\begin{keyword}
Ontologies \sep OWL \sep Context \sep Reification
\end{keyword}

\end{frontmatter}

\markboth{July 2016\hb}{July 2016\hb}

\section{Introduction}\label{sec:intro}

In knowledge representation, it is often necessary to characterize the context associated to a statement, such as when and how it was generated, or who uttered it. However, RDF and OWL can only represent natively binary relations~\cite{nardi_introduction_2003}. Many models exist for providing statements about statements, some of which are specific to a certain ``dimension'' of context, such as temporal validity.

Along these lines, in 2006, \citet{welty_reusable_2006} proposed an ontology for describing fluents (\ie entities whose characteristics change over time). Their approach advanced the state of the art in temporal representation, and has been used and extended in other works, but it only addresses one dimension characterizing a statement. Nonetheless, \citeauthor{welty_reusable_2006}'s approach can be extended so that any number of context's dimensions can be represented. We propose this extension through a generic ontology that can be extended to implement any specific dimension of context. In addition, we address the problem of modeling contextual datatype properties, as well as combining different contextual dimensions. This work is motivated by the need to characterize web datasets in terms of trust, provenance, and temporal validity, in the context of a question answering system within the WDAqua project.\footnote{\url{http://wdaqua.informatik.uni-bonn.de}}

In reference to the original 4dFluents ontology, we call our ontology NdFluents. We recall \citeauthor{welty_reusable_2006}'s contribution in \refsec{weltyfikes}, then we present the NdFluents ontology in \refsec{ourwork} together with some issues. In \refsec{practice}, we summarize guidelines for expressing contextualized statements, and apply these guidelines to a concrete use case that we published according to the Web of Data best practices. We then compare our approach to related work on representing the context of information in \refsec{relwork} before concluding.

\section{4dFluents Ontology}\label{sec:weltyfikes}

\citet*{welty_reusable_2006} address the problem of representing \emph{fluents}, \ie relations that hold within a certain time interval and not in others. They address the issue from the perspective of diachronic identity (that is, how an entity looks to be different at different times), showcasing the two different ways of tackling it:

\begin{itemize}
    \item The \emph{endurantist} (\textit{3D}) view maintains a differentiation between \emph{endurants}, entities that are present at all times during its whole existence, and \emph{perdurants}, events affecting an entity during a definite period of time during the entity's existence.
    \item The \emph{perdurantist} (\textit{4D}) view argues that entities themselves have to be handled as perdurants, \ie temporal parts of a four dimensional meta-entity. Instead of making an assertion about some entities, such as \emph{``Paris is the capital of France''}, one should make the assertion about their temporal parts: \emph{``A temporal part of Paris (since 508 up to now) is the capital of a temporal part of France (since 508 up to now)''}.
\end{itemize}

\citeauthor{welty_reusable_2006} adopt the perdurantist approach to create the \emph{4dFluents} ontology, representing \emph{entities at a time} and using them as resources for their statements. The 4dFluents ontology expressed in OWL2 Functional Syntax is shown in \refont{4dFluents}.

\begin{Ontology}
    \begin{lstlisting}
Prefix( 4d:=<http://www.example.com/4dFluents#> )
Ontology( <http://www.example.com/4dFluents>
	Declaration( Class( 4d:Interval ) )
	Declaration( Class( 4d:TemporalPart ) )
	DisjointClasses( 4d:Interval 4d:TemporalPart )
	
	Declaration( ObjectProperty( 4d:fluentProperty ) )
	ObjectPropertyDomain( 4d:fluentProperty 4d:TemporalPart )
	ObjectPropertyRange( 4d:fluentProperty 4d:TemporalPart )
	
	Declaration( ObjectProperty( 4d:temporalExtent ) )
	FunctionalObjectProperty( 4d:temporalExtent )
	ObjectPropertyDomain( 4d:temporalExtent 4d:TemporalPart )
	ObjectPropertyRange( 4d:temporalExtent 4d:Interval )
	
	Declaration( ObjectProperty( 4d:temporalPartOf ) )
	FunctionalObjectProperty( 4d:temporalPartOf )
	ObjectPropertyDomain( 4d:temporalPartOf 4d:TemporalPart )
	ObjectPropertyRange( 4d:temporalPartOf ObjectComplementOf( 4d:Interval ))
)
    \end{lstlisting}
    \vspace*{-7mm}
    \caption{4dFluents ontology (from \cite{welty_reusable_2006})}
    \label{ont:4dFluents}
\end{Ontology}

In order to use the ontology for describing fluents, one has to introduce axioms at the terminological level (TBox) as well as assertions in the knowledge base (ABox). For instance, if one wants to say that \emph{``Paris is the capital of France''} since 508, the relation ``capital of'' has to be a subproperty of \texttt{fluentProperty} and new individuals have to be introduced for the temporal part of Paris and of France, as shown in \refont{fluentstmt}.

\begin{Ontology}
    \begin{lstlisting}
Declaration( ObjectProperty( ex:capitalOf ) )
SubObjectPropertyOf( ex:capitalOf 4d:fluentProperty )
ClassAssertion( 4d:TermporalPart ex:Paris@508 )
ClassAssertion( 4d:TermporalPart ex:France@508 )
ClassAssertion( 4d:Interval ex:year508) )
ObjectPropertyAssertion( ex:capitalOf ex:Paris@508 ex:France@508 )
ObjectPropertyAssertion( 4d:temporalExtent ex:Paris@508 ex:year508 )
ObjectPropertyAssertion( 4d:temporalExtent ex:France@508 ex:year508 )
ObjectPropertyAssertion( 4d:temporalPartOf ex:Paris@508 ex:Paris )
ObjectPropertyAssertion( 4d:temporalPartOf ex:France@508 ex:France )
    \end{lstlisting}
    \vspace*{-7mm}
    \caption{Expressing a fact about a fluent entity with the 4dFluents ontology}\label{ont:fluentstmt}
\end{Ontology}

Welty and Fikes argue that, although there are various other ways of modelling fluents, the 4dFluents approach has proved being efficient in projects led by the authors. Based on these observations, we in turn think that adopting a similar approach, generalized to any dimension of context, would be a sensible choice.

\section{Extension of 4dFluents Ontology for Multiple Dimensions}\label{sec:ourwork}

In this section we discuss how to broaden the 4dFluents ontology to other dimensions different than time. In the first subsection we propose our ontology, then we proceed to discuss different representation details in the following subsections.

\subsection{Extending the 4dFluents Ontology}\label{ssec:ndfluents}

A temporal part of an entity can be viewed as an individual context dimension of the entity. A similar approach can then be used to represent different dimensions, such as provenance or confidence. Continuing with our running example, if Wikipedia states that \emph{``Paris is the capital of France''}, we can articulate that fact as \emph{``A Paris as defined by Wikipedia is the capital of France as defined by Wikipedia''}. Different context dimensions of an entity could then be combined if applicable, allowing to represent complex information, such as: \emph{``A temporal part Paris as defined by Wikipedia is the capital of a temporal part of France as defined by Wikipedia''}.

We use this idea to extend the 4dFluents ontology for any context dimension in the \emph{NdFluents} ontology. The ontology, shown in \refont{NdFluents}, and published in \url{http://www.emse.fr/~zimmermann/ndfluents.html}, is a direct extension from temporal parts to contextual parts.

% \begin{verbatim}
% Ontology(NdFluents
%     Class(Context partial)
%     Class(ContextualPart partial)
%     DisjointClasses(Context ContextualPart)
%     ObjectProperty(contextualProperty
%         domain(ContextualPart)
%         range(ContextualPart))
%     DatatypeProperty(contextualProperty
%         domain(ContextualPart))
%     ObjectProperty(contextualExtent Functional
%         domain(ContextualPart)
%         range(Context))
%     ObjectProperty(contextualPartOf Functional
%         inverseOf(hasContextualPart)
%         domain(ContextualPart)
%         range(complementOf(Context))))
% \end{verbatim}

\begin{Ontology}%[htbp]
    \begin{lstlisting}
Prefix( nd:=<http://purl.org/NET/NdFluents#> )
Ontology( <http://purl.org/NET/NdFluents>
	Declaration( Class( nd:Context ) )            
	Declaration( Class( nd:ContextualPart ) )     
	DisjointClasses( nd:Context nd:ContextualPart )

	Declaration( ObjectProperty( nd:contextualProperty ) )
	ObjectPropertyDomain( nd:contextualProperty nd:ContextualPart )
	ObjectPropertyRange( nd:contextualProperty nd:ContextualPart )

	Declaration( ObjectProperty( nd:contextualExtent ) )
	ObjectPropertyDomain( nd:contextualExtent nd:ContextualPart )
	ObjectPropertyRange( nd:contextualExtent nd:Context )

	Declaration( ObjectProperty( nd:contextualPartOf ) )
	FunctionalObjectProperty( nd:contextualPartOf )
	ObjectPropertyDomain( nd:contextualPartOf nd:ContextualPart )
	ObjectPropertyRange( nd:contextualPartOf ObjectComplementOf( nd:Context ))
)
    \end{lstlisting}
    \vspace*{-7mm}
    \caption{NdFluents ontology}
    \label{ont:NdFluents}
\end{Ontology}

Note that \texttt{FunctionalObjectProperty( nd:contextualExtent )} axiom is not included. This axiom should appear if the ontology was a direct translation from temporal dimension to a generic context dimension, but it is no longer applicable in the general case when we have more than one context dimension. Depending on the model used to represent the information, it will be necessary to add it (see \refssec{dimensions}).

The NdFluents ontology is meant to be implemented for different context dimensions in a modular way. In this sense, the 4dFluents ontology can be seen as a concrete implementation of NdFluents, as we show in \refont{4dFluentsUsingNdFluents}. In \reffig{NdFluents} we show the representation of a statement with temporal context using this ontology. The non-dashed parts are equivalent to the original 4dFluents ontology, while the dashed parts correspond to the NdFluents extension. Other context dimensions, such as provenance, can be modeled similarly to the temporal dimension by replacing \texttt{TemporalPart} with \texttt{ProvenancePart}, \texttt{temporalExtent} with \texttt{provenanceExtent}, \texttt{Interval} with \texttt{Provenance}, and \texttt{temporalPartOf} with \texttt{provenancePartOf}. Additionally, an assertion like \emph{``Paris is the capital of France, according to Wikipedia''} can be modeled following the same pattern as in \refont{fluentstmt}, replacing the property and class names with their counterpart in the provenance dimension.

\begin{Ontology}%[htbp]
    \begin{lstlisting}
Prefix( nd:=<http://purl.org/NET/ndfluents#> )
Prefix( 4d:=<http://purl.org/NET/ndfluents/4dFluents#>)
Ontology( <http://www.example.com/4dFluentsV2>
	Import( <http://www.example.com/NdFluents> )
	
	Declaration( Class( 4d:Interval ) )
	SubClassOf( 4d:Interval nd:Context )
	Declaration( Class( 4d:TemporalPart ) )
	SubClassOf( 4d:TemporalPart nd:ContextualPart )
	
	Declaration( ObjectProperty( :temporalExtent ) )
	SubObjectPropertyOf( 4d:temporalExtent nd:contextualExtent )
	ObjectPropertyDomain( 4d:temporalExtent 4d:TemporalPart )
	ObjectPropertyRange( 4d:temporalExtent 4d:Interval )
	
	Declaration( ObjectProperty( :temporalPartOf ) )
	SubObjectPropertyOf( 4d:temporalExtent nd:contextualPartOf )
	ObjectPropertyDomain( 4d:temporalPartOf 4d:TemporalPart )
)
    \end{lstlisting}
    \vspace*{-7mm}
    \caption{4dFluents ontology as implementation of NdFluents}
    \label{ont:4dFluentsUsingNdFluents}
\end{Ontology}

\begin{figure}%[htbp]
    \centering
    \scalebox{0.75}{%
    \begin{tikzpicture}[node distance = 1]
        \node[class,new] (ContextualPart) at (0,0) {ContextualPart};
        \node[class,below = of ContextualPart] (TemporalPart) {TemporalPart};
        \node[class,new,right = of TemporalPart] (ProvenancePart) {ProvenancePart};
        \node[class,new,right = of ProvenancePart] (TrustPart) {TrustPart};
        \path[new] (TemporalPart) to node {$\cdots$} (ProvenancePart);
        \path[new] (ProvenancePart) to node {$\cdots$} (TrustPart);
        \node[entity,below left = of TemporalPart] (Paris@1) {Paris@1};
        \node[entity,below right = of TemporalPart] (France@1) {France@1};
        \node[class,new,below = of TemporalPart, yshift = -1cm] (contextualExtent) {contextualExtent};
        \node[class,new,below = of contextualExtent] (Context) {Context};
        \node[class,below = of Context] (Interval) {Interval};
        \node[entity,below = of Interval] (t1) {$t_1$};
        \node[class,new,below = of t1] (contextualPartOf) {contextualPartOf};
%        \node[class,left = of contextualPartOf] (City) {City};
%        \node[class,right = of contextualPartOf] (Country) {Country};
        \node[entity,left = of contextualPartOf] (Paris) {Paris};
        \node[entity,right = of contextualPartOf] (France) {France};
        \path[type] (Paris@1) -- (TemporalPart);
        \path[type] (France@1) -- (TemporalPart);
        \path[type] (t1) -- (Interval);
%        \path[type] (Paris) -- (City);
%        \path[type] (France) -- (Country);
        \path[subclass,new] (TemporalPart) -- (ContextualPart);
        \path[subclass,new] (ProvenancePart) -- (ContextualPart);
        \path[subclass,new] (TrustPart) -- (ContextualPart);
        \path[subclass,new] (Interval) -- (Context);
        \path[property] (Paris@1) to node[fill=white] {capitalOf} (France@1);
        \path[property,bend right=45] (Paris@1) to node[fill=white] (temporalExtent1) {temporalExtent} (t1);
        \path[property,bend left=45] (France@1) to node[fill=white] (temporalExtent2) {temporalExtent} (t1);
        \path[property,bend right=45] (Paris@1) to node[fill=white, near end] (temporalPartOf1) {temporalPartOf} (Paris);
        \path[property,bend left=45] (France@1) to node[fill=white, near end] (temporalPartOf2) {temporalPartOf} (France);
        \path[subproperty,new] (temporalExtent1) -- (contextualExtent);
        \path[subproperty,new] (temporalExtent2) -- (contextualExtent);
        \path[subproperty,new] (temporalPartOf1) -- (contextualPartOf);
        \path[subproperty,new] (temporalPartOf2) -- (contextualPartOf);
    \end{tikzpicture}
    }
    \caption{Example of 4dFluents as implementation of NdFluents}
    \label{fig:NdFluents}
\end{figure}

%NdFluents opens a range of possibilities for representing different contexts of information about an entity. In the following sections, we showcase different scenarios where NdFluents can be useful, and discuss potential implementations of the ontology.

\subsection{Dealing with Datatype Properties}\label{ssec:datatypes}

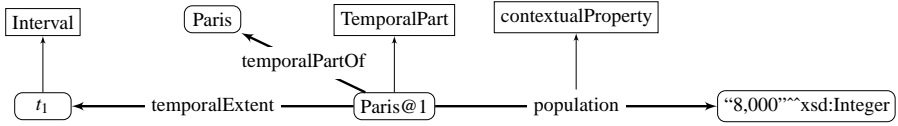
\begin{figure}%[htbp]
    \centering
    \scalebox{0.75}{%
    \begin{tikzpicture}[node distance = 1]
        \node[class] (TemporalPart) at (0,0)  {TemporalPart};
        \node[entity,below = of TemporalPart] (Paris@1) {Paris@1};
        \node[entity,right = of Paris@1,xshift=4cm] (Population) {``8,000''\^{}\^{}xsd:Integer};
        \node[entity,left = of Paris@1, xshift=-4cm] (t1) {$t_1$};
        \node[class,above = of t1] (Interval) {Interval};
%        \node[class,below left = of Paris@1] (City) {City};
        
        \path[type] (Paris@1) -- (TemporalPart);
        \path[property] (Paris@1) to node[fill=white] (population) {population} (Population);
        \node[class,above = of population] (contextualProperty) {contextualProperty};
        \path[type] (t1) -- (Interval);
%        \path[type] (Paris) -- (City);
        \path[property] (Paris@1) to node[fill=white] (temporalExtent) {temporalExtent} (t1);
        \node[entity,above = of temporalExtent] (Paris) {Paris};
        \path[property] (Paris@1) to node[fill=white] {temporalPartOf} (Paris);
        \path[subproperty] (population) -- (contextualProperty);
    \end{tikzpicture}
    }
    \caption{Example of Contextual Datatype Property}
    \label{fig:NdFluentsDatatypeProperties}
\end{figure}

The original 4dFluents ontology does not provide any information for modelling datatype properties. While there is nothing that prevents using regular datatype properties with contextual parts of an entity, it may be desirable to declare explicit axioms for context properties to facilitate reasoning on that information. In that case, the statements of \refont{NdFluentsDatatypeProperties} need to be added to the NdFluents ontology. \reffig{NdFluentsDatatypeProperties} shows an example where a contextual property is used to state the population of Paris in a specific temporal interval. Note that it is possible to create specific contextualProperty subproperties for different contextual dimensions (\ie \texttt{temporalProperty} for \texttt{TemporalPart}) for properties related to concrete context dimensions. 

\begin{Ontology}%[htbp]
    \begin{lstlisting}
Prefix( nd:=<http://purl.org/NET/ndfluents#> )
Ontology( <http://purl.org/NET/ndfluents/contextualDatatypeProperty>
	Declaration( DataProperty( nd:contextualDatatypeProperty ) )
	DataPropertyDomain ( nd:contextDataProperty nd:ContextualPart )
)
    \end{lstlisting}
    \vspace*{-7mm}
    \caption{Datatype axioms for NdFluents ontology}
    \label{ont:NdFluentsDatatypeProperties}
\end{Ontology}

\subsection{Combining Different Context Dimensions}\label{ssec:dimensions}

An important scenario where NdFluents becomes relevant is when the necessity of combining two or more dimensions of context arises, such as saying that \emph{``according to Wikipedia, Paris is the capital of France since 508''}. The NdFluents ontology supports different ways of representing the information. In this section we describe the most relevant models that can be used.

\paragraph{\textbf{Contexts in Context.}}

One possible model to represent information using different context dimensions is to relate a Contextual Part to another Contextual Part. This approach can be taken when the ``first level'' Contextual Parts  are relevant facts of the knowledge base, and the intention is to state additional information about them. To be able to reason about different contextual levels of any entity, it is desirable for the \texttt{contextualPartOf} property to be transitive, which can be achieved by adding the axiom of \refont{NdFluentsTransitive}.

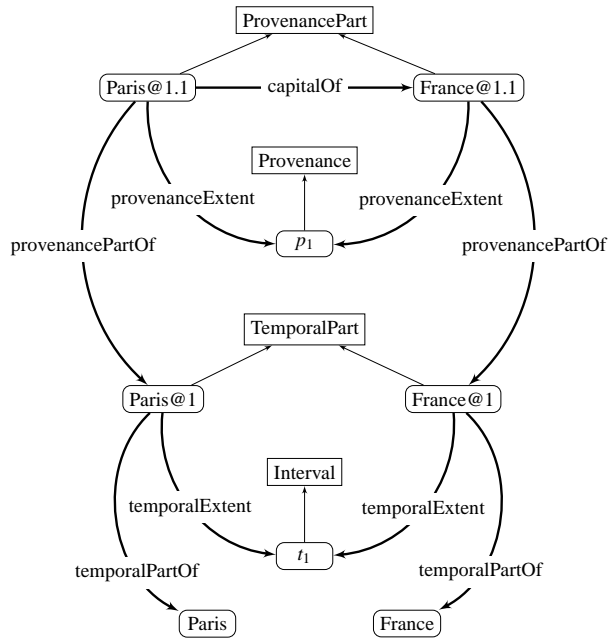
\begin{figure}%[htbp]
    \centering
    \scalebox{0.75}{%
    \begin{tikzpicture}[node distance = 1]
        \node[class,below] (ProvenancePart) at (0,0)  {ProvenancePart};
        \node[entity,below left = of ProvenancePart] (Paris@p1) {Paris@1.1};
        \node[entity,below right = of ProvenancePart] (France@p1) {France@1.1};
        \node[class,below = of ProvenancePart, yshift = -1cm] (Provenance) {Provenance};
        \node[entity,below = of Provenance] (p1) {$p_1$};
        
        \node[class,below = of p1] (TemporalPart) {TemporalPart};
        \node[entity,below left = of TemporalPart] (Paris@1) {Paris@1};
        \node[entity,below right = of TemporalPart] (France@1) {France@1};
        \node[class,below = of TemporalPart,yshift=-1cm] (Interval) {Interval};
        \node[entity,below = of Interval] (t1) {$t_1$};
%        \node[class,below left = of t1] (City2) {City};
%        \node[class,below right = of t1] (Country2) {Country};
        \node[entity,below left = of t1] (Paris2) {Paris};
        \node[entity,below right = of t1] (France2) {France};
        
        \path[type] (Paris@p1) -- (ProvenancePart);
        \path[type] (France@p1) -- (ProvenancePart);
        \path[type] (p1) -- (Provenance);
        \path[property] (Paris@p1) to node[fill=white] {capitalOf} (France@p1);
        \path[property,bend right=45] (Paris@p1) to node[fill=white] {provenanceExtent} (p1);
        \path[property,bend left=45] (France@p1) to node[fill=white] {provenanceExtent} (p1);
        \path[property,bend right=45] (Paris@p1) to node[fill=white] {provenancePartOf} (Paris@1);
        \path[property,bend left=45] (France@p1) to node[fill=white] {provenancePartOf} (France@1);
        
        \path[type] (Paris@1) -- (TemporalPart);
        \path[type] (France@1) -- (TemporalPart);
        \path[type] (t1) -- (Interval);
%        \path[type] (Paris2) -- (City2);
%        \path[type] (France2) -- (Country2);
        %\path[property] (Paris@1) to node[fill=white] {capitalOf} (France@1);
        \path[property,bend right=45] (Paris@1) to node[fill=white] {temporalExtent} (t1);
        \path[property,bend left=45] (France@1) to node[fill=white] {temporalExtent} (t1);
        \path[property,bend right=55] (Paris@1) to node[fill=white,near end] {temporalPartOf} (Paris2);
        \path[property,bend left=55] (France@1) to node[fill=white,near end] {temporalPartOf} (France2);
    \end{tikzpicture}
    }
    \caption{Contexts in Context}
    \label{fig:ContextsInContext}
\end{figure}

While data about different context dimensions can be more fine-grained using this model, it also grows in complexity. For example, in \reffig{ContextsInContext} the statement \texttt{capitalOf} is related to the provenance dimension Contextual Part \texttt{Paris@1.1}. This information is in no way related  to the Temporal Part \texttt{Paris@1}. While we could have this statement duplicated in the example, this is not possible as soon as we add another provenance Contextual Part to \texttt{Paris@1.1}. We believe that this model can be useful in some specific cases, but it is usually too cumbersome.

\begin{Ontology}%[htbp]
    \begin{lstlisting}
Prefix( nd:=<http://purl.org/NET/ndfluents#> )
Ontology( <http://purl.org/NET/ndfluents/transitivecontextualpartof>
	TransitiveObjectProperty( nd:contextualPartOf )
)
    \end{lstlisting}
    \vspace*{-7mm}
    \caption{Transitive axiom for NdFluents ontology}
    \label{ont:NdFluentsTransitive}
\end{Ontology}

\paragraph{\textbf{Use Multiple Contexts for each Contextual Part.}}

\begin{figure}%[htbp]
    \centering
    \scalebox{0.75}{%
    \begin{tikzpicture}[node distance = 1]
        \node (zero) at (0,0) {};
        \node[class,left = of zero] (TemporalPart) {TemporalPart};
        \node[class,right = of zero] (ProvenancePart) {ProvenancePart};
        \node[entity,below left = of TemporalPart] (Paris@1) {Paris@1};
        \node[entity,below right = of ProvenancePart] (France@1) {France@1};
        \node[class,below right = of Paris@1] (Interval) {Interval};
        \node[class,below left= of France@1] (Provenance) {Provenance};
        \node[entity,below = of Interval] (t1) {$t_1$};
        \node[entity,below = of Provenance] (p1) {$p_1$};
%        \node[class,below = of t1] (City) {City};
%        \node[class,below= of p1] (Country) {Country};
        \node[entity,below = of t1] (Paris) {Paris};
        \node[entity,below = of p1] (France) {France};
        \path[type] (Paris@1) -- (TemporalPart);
        \path[type] (France@1) -- (TemporalPart);
        \path[type] (Paris@1) -- (ProvenancePart);
        \path[type] (France@1) -- (ProvenancePart);
        \path[type] (t1) -- (Interval);
        \path[type] (p1) -- (Provenance);
%        \path[type] (Paris) -- (City);
%        \path[type] (France) -- (Country);
        \path[property] (Paris@1) to node[fill=white] {capitalOf} (France@1);
        \path[property,bend right] (Paris@1) to node[fill=white,pos="0.65"] {temporalExtent} (t1);
        \path[property,bend right=10] (France@1) to node[fill=white,near start] {temporalExtent} (t1);
        \path[property,bend left=10] (Paris@1) to node[fill=white,near start] {provenanceExtent} (p1);
        \path[property,bend left] (France@1) to node[fill=white,pos="0.65"] {provenanceExtent} (p1);
        \path[property,bend right=60] (Paris@1) to node[fill=white,pos="0.70",right] {provenancePartOf} (Paris);
        \path[property,bend left=60] (France@1) to node[fill=white,pos="0.70",left] {provenancePartOf} (France);
        \path[property,bend right=75] (Paris@1) to node[fill=white,near end,left] {temporalPartOf} (Paris);
        \path[property,bend left=75] (France@1) to node[fill=white,near end,right] {temporalPartOf} (France);
    \end{tikzpicture}
    }
    \caption{Multiple contexts on one ContextualPart}
    \label{fig:2contexts}
\end{figure}
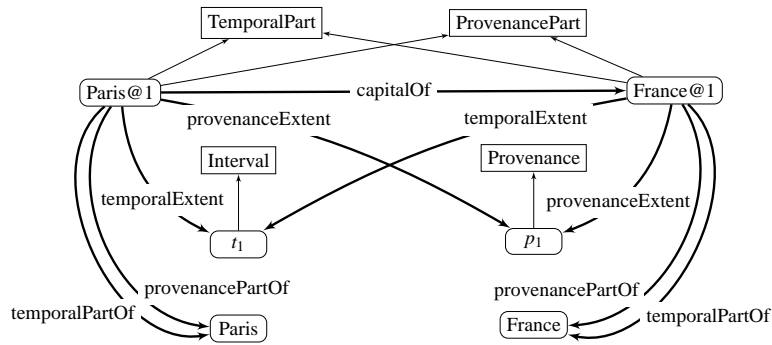

A more generic approach for representing entities with more than one contextual dimension is to have Contextual Parts with more than one Contextual Extent. Using this model, only one Contextual Part is created for a combination of context dimensions. This Contextual Part is then related to all the related contextual information, as shown in \reffig{2contexts}. This model is easier to model: Relating the Contextual Part with the context dimensions is straightforward. It also avoids ambiguity when modelling contextual information related to more than one contextual dimension, and reduces the number of resources in the ontology (\ie while the previous model needed one Contextual Part for each context dimension involved, this approach only requires one Contextual Part). Note that \texttt{contextualPartOf} is a functional property, which means that there cannot be a Contextual Part of more than one entity.

\paragraph{\textbf{Combine Different Contexts on one Contextual Extent.}}

Finally, a third possibility is to create Contextual Extents that combine two or more context dimensions, and enforce a limit of only one Contextual Extent per Contextual Part. This model adds a layer of complexity to the previous approach, but it can be useful to require a specific combination of context dimensions on a set of Contextual Parts. This can be achieved by adding the axiom in \refont{NdFluentsFunctional}.

%In addition, we can also specify that two different Contextual Parts with the same Contextual Extent by adding the axiom in \refont{NdFluentsInverseFunctional}. Note that while this axiom could be added to the previous approach, it would cause undesirable effects if Contextual Parts with different Contexts relate to the same Contextual Extent (\ie if a Contextual part of Paris which is at the same time Temporal Part and Provenance Part has a the same Temporal Extent as another Contextual Part that is only a Temporal part, both would be entailed to be the same).

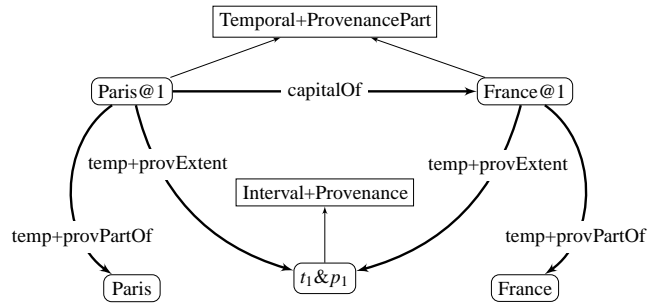
\begin{figure}%[htbp]
    \centering
    \scalebox{0.75}{%
    \begin{tikzpicture}[node distance = 1]
        \node[class] (CombinedPart) at (0,0) {Temporal+ProvenancePart};
        \node[entity,below left = of CombinedPart] (Paris@1) {Paris@1};
        \node[entity,below right = of CombinedPart] (France@1) {France@1};
        \node[class,below = of CombinedPart,yshift=-1.5cm] (CombinedContext) {Interval+Provenance};
        \node[entity,below = of CombinedContext] (tp1) {$t_1\&p_1$};
%        \node[class,below = of Paris@1,yshift=-2cm] (City) {City};
%        \node[class,below = of France@1,yshift=-2cm] (Country) {Country};
        \node[entity,below = of Paris@1,yshift=-2cm] (Paris) {Paris};
        \node[entity,below = of France@1,yshift=-2cm] (France) {France};
        \path[type] (Paris@1) -- (CombinedPart);
        \path[type] (France@1) -- (CombinedPart);
        \path[type] (tp1) -- (CombinedContext);
%        \path[type] (Paris) -- (City);
%        \path[type] (France) -- (Country);
        \path[property] (Paris@1) to node[fill=white] {capitalOf} (France@1);
        \path[property,bend right] (Paris@1) to node[fill=white,near start] {temp+provExtent} (tp1);
        \path[property,bend left] (France@1) to node[fill=white,near start] {temp+provExtent} (tp1);
        \path[property,bend right=55] (Paris@1) to node[fill=white,near end] {temp+provPartOf} (Paris);
        \path[property,bend left=55] (France@1) to node[fill=white,near end] {temp+provPartOf} (France);
    \end{tikzpicture}
    }
    \caption{Combination of different contexts on one contextual extent}
    \label{fig:CombineContexts}
\end{figure}

We show an example of this approach on \reffig{CombineContexts}. Note that the combined classes and properties are subclasses and subproperties of the corresponding classes and properties of the two context dimensions they are combining (\eg \texttt{Temporal+ProvenancePart} is subclass of \texttt{TemporalPart} and \texttt{ProvenancePart}). As a result, querying and reasoning can be performed in an identical way as the previous approach.

\begin{Ontology}%[htbp]
    \begin{lstlisting}
Prefix( nd:=<http://purl.org/NET/ndfluents#> )
Ontology( <http://purl.org/NET/ndfluents/functionalcontextualextent>
	FunctionalObjectProperty( nd:contextualExtent )
)
    \end{lstlisting}
    \vspace*{-7mm}
    \caption{Functional Contextual Extents axiom for NdFluents ontology}
    \label{ont:NdFluentsFunctional}
\end{Ontology}

%\begin{Ontology}%[htbp]
%    \begin{lstlisting}
%        InverseFunctionalObjectProperty( nd:contextualExtent )
%    \end{lstlisting}
%    \caption{Functional Contextual Extents axiom for NdFluents ontology}
%    \label{ont:NdFluentsInverseFunctional}
%\end{Ontology}

\subsection{Relations between Different ContextualParts}\label{ssec:relations}

The NdFluents ontology presented thus far allows to model relations among different contextual parts of different context dimensions (\ie a Temporal Part of Paris could be the capital of a Provenance Part of France). While this can be convenient for individual cases, it is often needed for a contextual property to be related to Contextual Parts of the same dimension of context. In this case, it is necessary to add the appropriate axioms to the ontology. In \refont{4dFluentsTemporalRestriction} we show the needed axioms to include this restriction on the Temporal Parts. Conversely, if there are datatype properties related to specific dimensions, axioms from \refont{4dFluentsDataTypeTemporalRestriction} should be added.

\begin{Ontology}%[htbp]
    \begin{lstlisting}
Prefix( nd:=<http://purl.org/NET/ndfluents#> )
Prefix( 4d:=<http://purl.org/NET/ndfluents/4dFluents#>)
Ontology( <http://purl.org/NET/ndfluents/4dFluents/temporalpartrestriction>
	Declaration( ObjectProperty( 4d:fluentProperty ) )
	SubObjectPropertyOf( 4d:fluentProperty nd:contextualProperty )
	ObjectPropertyDomain( 4d:fluentProperty 4d:TemporalPart )
	ObjectPropertyRange( 4d:fluentProperty 4d:TemporalPart )
)
    \end{lstlisting}
    \vspace*{-7mm}
    \caption{Temporal restriction on object properties 4dFluents ontology}
    \label{ont:4dFluentsTemporalRestriction}
\end{Ontology}

\begin{Ontology}%[htbp]
    \begin{lstlisting}
Prefix( nd:=<http://purl.org/NET/ndfluents#> )
Prefix( 4d:=<http://purl.org/NET/ndfluents/4dFluents#>)
Ontology( <http://purl.org/NET/ndfluents/4dFluents/temporalpartrestriction>
	Declaration( DataProperty( 4d:fluentDataTypeProperty ) )
	SubDataPropertyOf( 4d:fluentDataTypeProperty nd:contextualProperty )
	DataPropertyDomain( 4d:fluentProperty 4d:TemporalPart)
)
    \end{lstlisting}
    \vspace*{-7mm}
    \caption{Temporal restriction on datatype properties 4dFluents ontology}
    \label{ont:4dFluentsDataTypeTemporalRestriction}
\end{Ontology}

In a similar fashion, it is usually desirable that Contextual Parts of the same contextual dimension relate to the same Contextual Extent. That is, if a Provenance Part of Paris relates to a Provenance Part of France, their \texttt{provenanceExtent} properties should have the same Provenance object. However, this restriction cannot be expressed in OWL. If needed, a rule language (such as SWRL~\cite{horrocks_swrl:_2004} or RIF\footnote{\url{https://www.w3.org/TR/rif-overview}}) can be used for this purpose, but this case goes beyond the scope of this paper.

\subsection{Adapting Non-Contextual Ontologies to NdFluents}\label{ssec:withnoncontextual}

In the previous sections, we show how to define and use NdFluents to define properties that apply to contextual parts. However, in some cases, we would rather reuse existing ontologies. It is not always possible to add a subproperty relationship between an already defined property and a contextual property because most restrictions will not hold for the contextual parts of an entity. For instance, 
%the property \texttt{foaf:mbox\_sha1sum} in the FOAF ontology is an inverse functional datatype property\footnote{Inverse functional datatype properties are not allowed in OWL 2 DL, but the same reasoning could be applied to keys, that can be used as an approximation of inverse functional properties.}. If an email address is reused for different people (such as \texttt{azimmerm@liris.cnrs.fr} for both Antoine Zimmermann and Andreas Zimmermann\footnote{This is a real life example.}), then two distinct temporal parts of different persons would be inferred as the same entity.
let us suppose that the property \texttt{capitalOf} has domain \texttt{City}, and we use it as a fluent property in \reffig{NdFluents}. Then, it would be inferred that \texttt{Paris@1} is a city, instead of \texttt{Paris} being inferred as a city. There might be also a problem if an \texttt{InverseFunctionalObjectProperty} is used in two different contextual parts of an entity with the same object (\eg \texttt{Paris@508} and {Paris@2016} both being \texttt{capitalOf} \texttt{France}), because then they would be inferred to be the same entity.

To address this situation, instead of using the original properties, it is necessary to create new properties that are somehow related to the original. It is then necessary to model the new properties using Class Expressions for the restrictions. We show an example for the domain and range of the property \texttt{capitalOf} in \refont{RelatedContextualPropertExample}.

%\begin{Ontology}%[htbp]
%    \begin{lstlisting}
%Prefix( nd:=<http://purl.org/NET/ndfluents#> )
%Ontology( <http://purl.org/NET/ndfluents/4dFluents/related>
%	Declaration( ObjectProperty( nd:relatedProperty ) )
%	ObjectPropertyDomain( nd:relatedProperty nd:contextualProperty )
%	ObjectPropertyRange( nd:relatedProperty ObjectComplementOf( nd:contextualProperty ))
%)
%    \end{lstlisting}
%    \caption{Related Contextual Properties}
%    \label{ont:RelatedContextualProperty}
%\end{Ontology}
%\noteby{Jos\'e}{IS THIS EVEN POSSIBLE???}

\begin{Ontology}%[htbp]
    \begin{lstlisting}
Declaration( ObjectProperty( ex:contextualCapitalOf ) )
SubObjectPropertyOf( ex:contextualCapitalOf 4d:fluentProperty )
ObjectPropertyDomain( nd:contextualCapitalOf ObjectAllValuesFrom( ex:contextualPartOf ex:City) )
ObjectPropertyRange( nd:contextualCapitalOf ObjectAllValuesFrom( ex:contextualPartOf ex:Country ))
    \end{lstlisting}
    \vspace*{-7mm}
    \caption{Definition of Related Contextual Property to \texttt{capitalOf}}
    \label{ont:RelatedContextualPropertExample}
\end{Ontology}

\subsection{Dealing with Terminological Statements}\label{ssec:terminological}

In general, contexts are used just for assertions (what is usually considered the \emph{ABox}). However, there are cases where terminological statatements (the \emph{TBox}) can also be viewed under different context dimensions. Consider the case of classic biological kingdoms. While in the U.S. it has been traditionally considered that there are six kingdoms (\textit{Animalia, Plantae, Fungi, Protista, Archaea/Archaeabacteria, and Bacteria/Eubacteria}), many other countries consider only five (\textit{Animalia, Plantae, Fungi, Protista and Monera}). This classification evolved from the historic view of animal and plant kingdoms, while more recent classifications include up to eight kingdoms. Whether to include viruses in the taxonomy is still an ongoing debate.

Such classification would induce \texttt{subClassOf} relations such as \texttt{Haloarchaea subClassOf Archaeabacteria} and \texttt{Haloarchaea subClassOf Monera}, which hold for different temporal and provenance domains. However, to make these statements contextual, it is necessary to define a contextual subproperty related to \texttt{subClassOf}. This is possible, but it is important to take into account that the created property will not benefit from the standard inferences associated with subClassOf.

%There is no solution using only OWL that allows us to perform that kind of inference.

%It would not be possible to use the standard subClassOf relation as a contextual property using NdFluents in an OWL 2 DL ontology. While it is possible to create an OWL Full ontology in this case, it is not advisable due to the undecidability of OWL Full.

\section{NdFluents in Practice}\label{sec:practice}

In this section, we concretize the previous information on actual steps to implement the NdFluents ontology in practice, with a focus on the decisions to make in each step, and our recommendation. Then, we proceed to demonstrate an actual implementation of the NdFluents ontology for a concrete set of data with different context dimensions following those steps.

\subsection{Modeling a Knowledge Base}\label{ssec:guidelines}

In order to model a knowledge base with a number of context dimensions, it is necessary to model the ontology in the TBox, and then create the statements in the ABox using the ontology. The ontology can be modeled according to the following steps:

\begin{enumerate}
    \item For each context dimension, create the appropriate subclass of \texttt{ContextualPart}, and a subclass of \texttt{Context} (such as \texttt{TemporalPart} and \texttt{Interval} for temporal dimension).
    \item For each context dimension, create a subproperty of \texttt{contextualExtent} and \texttt{contextualPartOf} (such as \texttt{provenanceExtent} and \texttt{provenancePartOf} for Provenance dimension).
    \item If any contextual part includes datatype properties, it has to be decided whether to use the datatype restriction axioms (\refont{NdFluentsDatatypeProperties}). We advise to include them to improve reasoning capabilities.
    \item If there is more than one context dimension in the ontology, the model to represent the information needs to be selected (see \refssec{dimensions}). We recommend to use the second approach (\emph{use different Contexts for each Contextual Part}, see \reffig{2contexts}). If \emph{Contexts in Context} (\reffig{ContextsInContext}) is used, the transitivity axiom to the \texttt{contextualPartOf} property (\refont{NdFluentsTransitive}) needs to be added. If \emph{combine different contexts in Contextual Extents} (third approach) is used, create the combined subclasses of \texttt{ContextualPart} and \texttt{Context}, and the combined subproperties of \texttt{contextualExtent} and \texttt{contextualPartOf}, as shown in \reffig{CombineContexts}.
    \item If there is more than one context dimension, it is possible to use the same contextual properties for every dimension or create a different subproperty of \texttt{contextualProperty} for each one (see \refssec{relations}). For the last case, the context restrictions axioms (see \refont{4dFluentsTemporalRestriction} for temporal dimension) need to be added for each dimension. We recommend to include these axioms.
\end{enumerate}

Once  the  ontology  is  modeled,  the  next  series  of  steps  are  needed  to  model  the
contextual statements:

\begin{enumerate}
    \item For each context dimension, it is necessary to create the related Context information. This is a new resource of the related \texttt{Context} subclass and the adequate information (interval for temporal context, provenance information for provenance context, \etc). Depending on the model you choose to represent the information, it is needed to create a different resource for each context of an entity (Figures \ref{fig:ContextsInContext} and \ref{fig:2contexts}), or a unique resource that combines the contexts (\reffig{CombineContexts}).
    \item For each entity, create its related Contextual Parts. These are new resources with type the appropriate \texttt{ContextualPart} subclasses (\eg \texttt{TemporalPart}, \texttt{ProvenancePart}). These resources are connected to their non-contextual entity by \texttt{contextualPartOf} subproperties (\texttt{temporalPartOf, provenancePartOf}), and to the information related to the context, modeled as Context resources (\ie \texttt{Interval}, \texttt{Provenance}), by \texttt{contextualExtent} subproperties (\texttt{temporalExtent}, \texttt{provenanceExtent.}).
\end{enumerate}

In the following section, we present an example of a concrete use case where we follow those steps to model a knowledge base.

\subsection{Practical Use Case}\label{ssec:usecase}

In this section the NdFluents ontology is used on a practical example with two contextual dimensions: The estimated evolution of Earth population according to different sources, which needs temporal and provenance dimensions. For this task, we use the information provided by Wikipedia\footnote{\url{https://en.wikipedia.org/wiki/World_population_estimates}}. 

We model the TBox according to the steps defined in the previous section for the Temporal and Provenance dimensions. As the population of each period will be defined as datatype properties, we decide to include the datatype restriction axioms (\refont{NdFluentsDatatypeProperties} for temporal dimension, and similarly for the provenance dimension) in step 3. In step 4, we choose to \emph{use multiple Contexts for each Contextual Part} (\reffig{2contexts}) to model the information. The resulting TBox will be comprised of ontologies \ref{ont:NdFluents}, \ref{ont:4dFluentsUsingNdFluents}, \ref{ont:NdFluentsDatatypeProperties}, and the corresponding implementation of NdFluents for provenance (equivalent to \refont{4dFluentsUsingNdFluents}).

The contextual statements are also modeled following the steps of previous section. In step 1, we create the Intervals and the Provenance extents for every period and source. To model Intervals we use the OWL-Time ontology\footnote{\url{https://www.w3.org/TR/owl-time}}, while for Provenance Extents we use the PROV-O ontology\footnote{\url{https://www.w3.org/TR/prov-o}} (along with OWL-Time and the Event ontology\footnote{\url{http://motools.sourceforge.net/event/event.html}} to model the Activity). In step 2, we create the contextual parts corresponding to each period and source for the Earth. Those parts are of type \texttt{TemporalPart} and \texttt{ProvenancePart}, are defined as \texttt{temporalPartOf} and \texttt{provenancePartOf} of \texttt{Earth}, and are connected to their corresponding Intervals and Provenance Extents by \texttt{temporalExtent} and \texttt{provenanceExtent} properties. Finally, they include the population as a datatype property (note that in Wikipedia a few number of population values are given as an interval, in which case we use the average value).

The complete dataset is published in \url{http://www.emse.fr/~zimmermann/ndfluents.html}

With this data it is possible to make queries using SPARQL about specific periods and sources, obtaining individual or aggregated data. For example, it is possible to obtain the average population and number of studies per year using the \refqry{YearZero}. Using this result one could, for every period with at least two studies, compute the p-value Student's t-distribution considering the population of each study as theoretical mean. This value then could be attached to the contextual part as a new Trust Extent.

\begin{Query}
    \begin{lstlisting}
        SELECT ?year (AVG(?population) AS ?average) (COUNT(?earth_part) AS ?count)
        WHERE {
        	?earth_part nd:temporalExtent [
        	    time:intervalDuring [ 
        	        time:hasDateTimeDescription [ time:year ?year ]]] ;
        	    dbo:populationTotal ?population .
        }
        GROUP BY ?year
    \end{lstlisting}
    \vspace*{-7mm}
    \caption{Extracting statistical data for year zero}
    \label{qry:YearZero}
\end{Query}

\section{Related work}\label{sec:relwork}

In this section we describe other solutions to state information about statements. A number of them rely on ontology or modeling techniques, while some others modify or extend the underlying semantic technologies.

Descriptions and Situations ontology~\cite{gangemi_understanding_2003} is a work that precedes \citeauthor{welty_reusable_2006}'s that tries to describe ``contexts, methods, norms, theories, situations, and models at first-order, thus allowing a partial specification of those entities''. The descriptions represent conceptual elements (like laws, norms, regulations, crime types, \etc), while the situations represent observable elements (like legal facts,
cases, states of affairs, \etc)~\cite{gangemi_norms_2007}.

\citet{zamborlini_representation_2010} present an alternative work to 4dFluents, where they present two different alternatives to represent temporally changing information in OWL. Both approaches have a similar model to \citeauthor{welty_reusable_2006}'s, where the entities are sliced for different times. The main difference is that in the first one, \emph{Individual Concepts and Rigidity}, the original individuals are considered as classes. Thus, they are not described by any property, and a new slice has to be created every time that a property changes. On the other hand the second approach, ``Objects and Moments'', is based on \emph{Relators} and \emph{Qua-individuals}~\cite{masolo_relational_2005}, where the individuals are represented by an entity, and their slices inherit its properties. Then, any time a property changes, it is reflected in the original entity. The first approach is more prone to the proliferation of timeslices, and can only guarantee the immutability of original properties only by repetition on every timeslice. The second approach solves those issues at the cost of blurring the details of the changes of individual properties, and it is not clear how inheritance works in OWL.

In a later work \cite{zamborlini_ontologically-founded_2013}, \citeauthor{zamborlini_ontologically-founded_2013} focus on solving the issues of the prior approaches for representing events and properties of individuals. They maintain the fluent-like representation for events, but move to an N-ary representation (see below) for properties. However, they still not address the possibility to have more that one domain relation, nor address how inheritance is performed in OWL.

NdFluents is not the first extension of 4dFluents. There are a number of extension for Spatio-Temporal representations. \citeauthor{batsakis_representing_2011} \cite{batsakis_representing_2011,batsakisa_temporal_2011} enhance the 4dFluents mechanism  with qualitative temporal expressions to represent relations between intervals, and allow the definition of new intervals using this relations. Later, they extend it with several types of qualitative spatial relations and use it for their SOWL query language. \citeauthor{milea_towl:_2012} develop \emph{tOWL} \cite{milea_towl:_2012}, an extension of OWL for temporal domains, on top of a 4dFluents layer. \citeauthor{harbelot_continuum:_2013} \cite{harbelot_continuum:_2013,harbelot_spatio-temporal_2013} use tOWL and GeoSPARQL for spatio-temporal representations of entities. \citet{welty_context_2010} generalized later 4dFluents to Context Slices by changing the temporal part and context to contextual part and context, but does not make possible to use more than one context in the same dataset. To the best of our knowledge, NdFluents \emph{is} the first generic extension of 4dFluents for any number of arbitrary context dimensions.

\citet{krieger_temporal_2012}, instead of modeling temporal information with RDF, abandon the concept of triples and use quintuples to represent temporal information, where the 4th and the 5th elements represent the starting and ending instant of the event. However, this kind of solution requires tuples with an arbitrary number of elements.

There are other approaches to model arbitrary contextual information about entities or statements. Reification\footnote{\url{http://www.w3.org/TR/2014/REC-rdf11-mt-20140225/\#reification}} is the standard W3C model to represent information about an statement, proposed in 2004. A statement is represented as an instance of \texttt{rdf:statements}, which relates to the original triple with the properties \texttt{rdf:subject}, \texttt{rdf:predicate} and \texttt{rdf:object}. However, reification lacks formal semantics to connect the original triple with the reified statement, which disallows any reasoning on the information. N-Ary relations\footnote{\url{https://www.w3.org/TR/swbp-n-aryRelations}} were proposed in 2006 to represent relations between more than two individuals, or to describe the relation themselves. In this model, an individual is created to represent the relation, which can be used as the subject for new statements. While N-ary relation are an improvement over reification, it does not allow complete OWL inference. For instance, it is not possible to perform any reasoning involving inverse, transitive or symmetric relationships. Wikidata\footnote{\url{https://www.wikidata.org}} makes use of an specific implementation of N-ary relations, where each entity is related with statement, that in turn can be related to values, qualifiers, or references. The estimated word population, with temporal qualifiers for the date, and references for the provenance, is modeled using this pattern in Wikidata.\footnote{\url{https://www.wikidata.org/wiki/Q2\#P1082}} The Singleton Property \cite{nguyen_dont_2014,nguyen_reasoning_2015} is a recent proposal to represent information about statements in RDF. A particular instance of the predicate is created for every triple. This instance is related to the original predicate by the \texttt{singletonPropertyOf} property. Then, each statement can be unequivocally referenced using its predicate for attaching additional information. The Singleton Property is an intuitive approach, but it cannot be expressed in OWL and needs to extend the RDF semantics to make reasoning possible.

\citet{gangemi_multi-dimensional_2013} perform an empirical analysis of different patterns to represent metadata about statements in RDF and OWL. According to their results, N-ary relations and fluents have the higher reasoning capabilities and the possibility to add new arguments when needed, and also preserve the FOL relation topology. From those, however, only N-ary relations are intuitive, along with N-Quads. On the other hand N-Quads have the least impact in size while fluents have the higher. \citet{scheuermann_empirical_2013}, on the other side, perform a qualitative research that compares user preferences and ability for using different design patterns. In their study the fluents pattern is regarded as the most complicated and less used to model, while making a temporal slice of the predicate (which could be represented using the Singleton Property in RDF) seems more intuitive. The N-ary pattern is the model most frequently used. The model regarded as the most user-friendly is not representable using OWL, because it requires having a predicate as an argument of another (an approximation in RDF could be using N-Quads, though).

\section{Conclusion}\label{sec:ccl}

Representing contextual information in different dimensions is a current challenge in OWL. We have proposed NdFluents, a multi-domain contextual representation, based on the 4dFluents ontology. This is to the best of our knowledge, the first generic extension of 4dFluents for any number of arbitrary contextual dimensions. This representation is intended to be extended in a modular way for each desired context dimension in the ontology. We have discussed different possible models that can be used when combining different dimensions, possible additions to consider depending on the data we want to represent, and open issues of the ontology. We have also provided schematic guidelines for using NdFluents and a practical example. Both the ontology and the example are published for public usage. While NdFluents needs the inclusion of more triples and seems less intuitive than any of the alternatives, it allows for a more complete OWL inference within a context (with the exceptions presented in \refssec{withnoncontextual}). In addition NdFluents is the only approach where it is possible to add contextual information not only to an statement, but to an entity. The model also allows to retrieve easily all the information within a context for the same entity.

%Future work to consider is extending the ontology for the TBox and representing information that cannot be expressed with OWL, such as inferring that two contextual parts of the same entity and with the same contexts, which refer to the same contextual extent, are the same. It would also be relevant to study the possibility of mixing different approaches for representing contextual information (\ie using Singleton Properties and NdFluents on the same data, possibly for different context dimensions).

As future work, we want to apply this model to real world datasets. Our goal is to exploit the context of information to make the datasets fit for question answering, as well as, determine the most relevant data sources. This includes providing additional information based on the context and helping to find the most trustworthy data for the answer. A particular challenge lies in finding the temporal validity of the facts found in the data. 

\paragraph{Acknowledgement:}

This project is supported by funding received from the European Unions Horizon 2020 research and innovation program under the Marie Sk\l{}odowska-Curie grant agreement No 642795.

\bibliographystyle{plainnat}
\bibliography{main}

\end{document}